\documentclass{article}
\usepackage{ijcai19}

\usepackage{times}
\usepackage{soul}
\usepackage{url}
\usepackage[hidelinks]{hyperref}
\usepackage[utf8]{inputenc}
\usepackage[small]{caption}
\usepackage{graphicx}
\usepackage{amsmath}
\usepackage{booktabs}
\usepackage{algorithm}
\usepackage{algorithmic}
\usepackage{amssymb}
\usepackage{amsthm}
\usepackage{color}
\usepackage{enumitem}

\newtheorem{theorem}{Theorem}
\newtheorem{lemma}{Lemma}
\newtheorem*{remark}{Remark}

\title{MOORe: Model-based Offline-to-Online Reinforcement Learning}

\author{
Yihuan Mao$^1$\thanks{~~The work was done when the author visited Huawei Noah's Ark Lab.}
\and
Chao Wang$^2$\and
Bin Wang$^3$\And
Chongjie Zhang$^4$
\affiliations
$^{1,4}$Institute for Interdisciplinary Information Sciences, Tsinghua University\\
$^{2,3}$Huawei Noah's Ark Lab\\
\emails
maoyh20@mails.tsinghua.edu.cn,
\{wangchao358,wangbin158\}@huawei.com,
chongjie@tsinghua.edu.cn
}

\begin{document}

\maketitle

\begin{abstract}
With the success of offline reinforcement learning (RL), offline trained RL policies have the potential to be further improved when deployed online. A smooth transfer of the policy matters in safe real-world deployment. Besides, fast adaptation of the policy plays a vital role in practical online performance improvement. To tackle these challenges, we propose a simple yet efficient algorithm, \textbf{M}odel-based \textbf{O}ffline-to-\textbf{O}nline \textbf{Re}inforcement learning (MOORe), which employs a prioritized sampling scheme that can dynamically adjust the offline and online data for smooth and efficient online adaptation of the policy. We provide a theoretical foundation for our algorithms design. Experiment results on the D4RL benchmark show that our algorithm smoothly transfers from offline to online stages while enabling sample-efficient online adaption, and also significantly outperforms existing methods.

\end{abstract}


\section{Introduction}
\label{sec1}

Offline RL is a widely studied area that strongly connects to the real-world applications, as online data is expensive in many real environments \cite{healthcare,DBLP:journals/corr/abs-1805-04687}. Furthermore, when we have a well-trained offline agent, we can further enhance it by online interactions, which is the main concern of offline-to-online RL. A trivial transfer method may result in performance degradation, which is unacceptable since it may cause profit decline in recommendation systems, stock market trading, etc., or even severe safety issues in autonomous driving scenarios. Besides, if the transferred policy cannot quickly adapt to online settings, the offline-to-online transfer becomes useless.

Several methods have been proposed to tackle these challenges. For example, \cite{nair2021awac} improves the policy in a conservative manner in both offline and online stages, which fails to adapt to the online data, causing sample inefficiency. In contrast, \cite{lee2021offlinetoonline} introduces a balanced replay mechanism to boost transfer smoothness. However, significant performance drop can also be observed in some environments. \cite{anonymous2022adaptive} controls distribution shift by adaptively weighting the behavior cloning loss during online fine-tuning based on the agent’s performance and training stability, but it still cannot avoid performance degradation in some environments due to the difficulty of hyper-parameter tuning. Besides, all existing offline-to-online methods are model-free, which are of low data efficiency.

The failure of smooth offline-to-online transfer is mainly due to the ineffective control of distribution shift in existing methods. The offline data may be composed of transitions gathered by one or multiple unknown policies, while in the online training stage, transitions are gathered by the currently learned policy. To alleviate the distribution shift problem, the transferred policy should use more offline data to improve its performance instead. However, to quickly adapt to the online setting, the policy should care more about the newly collected online data.
It seems that the two purposes conflict to each other at the same time, since focusing on offline data would result in ignorance on the newest online data, and vice versa.

To address this dilemma in this paper, we propose a \textbf{M}odel-based \textbf{O}ffline-to-\textbf{O}nline \textbf{Re}inforcement learning (MOORe) algorithm, which consists of a simple yet efficient prioritized sampling scheme, along with the model-based learning framework. Specifically, the prioritized sampling scheme dynamically assigns different priority weights to offline and online transitions, aiming to control distribution shift during the offline-to-online learning stages. 
The data sampling scheme in our algorithm encourages smooth transfer in the early online stage, and fast adaptation in the late online stage. The integration of the prioritized sampling scheme and model-based framework reaches both smooth transfer and fast adaption, achieving the state-of-the-art performance in the offline-to-online RL setting on the D4RL benchmarks \cite{DBLP:journals/corr/abs-2004-07219}.

The main contributions of our method are two-fold:
\begin{itemize}
    \item We are the first to use model-based methods to effectively solve the offline-to-online RL problem, which significantly outperforms existing methods.
    \item Based on theoretical foundations, MOORe uses a prioritized sampling scheme to dynamically adjust the priority of offline and online data and achieves smooth transfer from the offline policy as well as fast adaption to the online data.
\end{itemize}


\section{Related works}\label{sec:related}
\paragraph{Offline RL} Different from offline-to-online RL, Offline RL focuses on training a policy using solely pre-existing datasets. Extrapolation error \cite{DBLP:journals/corr/abs-2007-08202,yu2020mopo} (a.k.a. bootstrapping error) and distribution shift \cite{fujimoto2018offpolicy} between offline dataset and online interactions are the main concerns, which are addressed by adding various constraints on the policy \cite{kumar2019stabilizing,wu2019behavior} or the Q values \cite{kumar2020conservative}.

Model-based methods are also used to learn conservative policies \cite{yu2020mopo,kidambi2020morel,yu2021combo}. MOPO \cite{yu2020mopo} adds penalty to the estimated reward function generated by an ensemble of estimated environment dynamics models to prevent the learned policy from visiting rare or unseen states. MoREL \cite{kidambi2020morel} takes another idea of making adjustments directly in the MDP, marking unseen states as fake terminal states.

\paragraph{Offline-to-online RL} 
In the offline-to-online RL scenario, \cite{nair2021awac} continues to improve the policy using online interactions while maintaining the conservative property. Researches afterwards employ techniques such as model ensemble and offline data re-sampling schemes \cite{lee2021offlinetoonline} to accelerate the online training procedure. \cite{anonymous2022adaptive} applies behavior cloning constraints to boost stability in the offline-to-online learning stage, while requiring a finely designed hyper-parameter tuning approach to adjust the constraint. Different from these methods, our model-based method employs a prioritized data sampling scheme to boost transfer smoothness and fast adaptation.

A similar but different area is Learning from Demonstrations (LfD) \cite{hester2017deep,reddy2019sqil}, which leverages offline dataset to accelerate the online training process \cite{pmlr-v80-kang18a}. Different from offline-to-online RL, LfD generally learns a policy from scratch until convergence, pre-existing data is used to speed up this process. Though a policy can be pre-trained before online interaction, LfD does not require the pre-trained policy to act well online.

\section{Method}
\label{sec:method}
In this section, we first present some technical preliminaries, and then introduce the details of our proposed method, followed by theoretical analyses.  

\subsection{Preliminaries}
\label{mboorl}
The standard RL framework assumes that the problem to be solved can be modeled as a Markov Decision Process (MDP) $M=(\mathcal{S},\mathcal{A},p,r,\mu_0,\gamma)$, where $\mathcal{S}$ and $\mathcal{A}$ denote the state and action spaces, $p(s'|s,a)$ is the transition dynamics and $r(s,a)$ is the reward function. $\mu_0$ is the distribution of initial states. $\gamma\in(0,1)$ is the discount factor, used to calculate the value function $V^\pi(s)=\mathbb{E}_{\pi}[\sum_{t=0}^\infty\gamma^tr(s_t,a_t)|s_0=s]$,
and the expected discounted return 
\begin{equation}\label{equ:target}
    \eta(\pi)=\mathbb{E}_{s\sim\mu_0}[V^\pi(s)]
\end{equation},
where $\pi(a|s)$ is a policy. The goal of RL is to find an optimal policy $\pi^*$ that maximizes the expected discounted return: $\pi^*=\mathop{\arg\min}_{\pi}\eta(\pi)$.

In Model-based RL, an environment dynamics model $\hat{M}$, consisting of the transition function $\hat{T}$ and the reward function $\hat{r}$, is estimated using a given dataset $\mathcal{D}$. The RL algorithm then learns a policy from the transitions generated by the learned model $(\mathcal{S},\mathcal{A},\hat{p},\hat{r},\mu_0,\gamma)$.

As an offline RL algorithm, MOPO\cite{yu2020mopo} learns an offline policy $\pi_{off}$ from a dataset $\mathcal{D}_{model}$ generated by an uncertainty-penalized MDP $\tilde{M}=(\mathcal{S},\mathcal{A},\hat{p},\tilde{r},\mu_0,\gamma)$, which is estimated from the offline dataset $\mathcal{D}_{off}$, where
\begin{equation}\label{equ:uncertainty_reward}
    \tilde{r}(s,a)=\hat{r}(s,a)-\lambda u(s,a),
\end{equation}
$\hat{p}(s, a)$ and $\hat{r}(s,a)$ are the estimated transition dynamics and reward function, respectively, and $u:\mathcal{S}\times\mathcal{A}\rightarrow\mathbb{R}$ is an uncertainty term describing the uncertainty of $(s,a)$, which is estimated by model ensemble techniques.

For brevity, we denote $\mathcal{D}_{on}$ as the dataset storing online interactions, $\pi_{on}$ as the policy learned from both $\mathcal{D}_{on}$ and $\mathcal{D}_{off}$, and $\hat{M}_{off} $ and $\hat{M}_{on}$ the environment dynamic models in the offline and online stages respectively.

\subsection{Model-based Offline-to-online RL}
\label{MOORe}

\begin{algorithm*}[t]
	\renewcommand{\algorithmicrequire}{\textbf{Input:}}
	\renewcommand{\algorithmicensure}{\textbf{Output:}}
	\caption{\textbf{M}odel-based \textbf{O}ffline-to-\textbf{O}nline \textbf{Re}inforcement learning (MOORe)}
	\label{alg1}
	\begin{algorithmic}[1]
	    \STATE \# \textbf{Offline learning stage}
		\STATE Run MOPO on  $\mathcal{D}_{off}$ to obtain $\pi_{off}$, an ensemble of $K$ estimated models $\{\hat{M}_{off}\}_{k=1}^K$, and the action-value functions $\hat{Q}_{off}$
		\STATE \# \textbf{Online learning stage}
		\STATE Initialize $\pi_{on} = \pi_{off}$, $\{\hat{M}_{on}=\hat{M}_{off}\}_{k=1}^K$, the action-value function $\hat{Q}_{on} = \hat{Q}_{off}$
		\STATE Initialize $\mathcal{D}_{on}=\emptyset$, $\mathcal{D}_{model}=\emptyset$ and the number of gradient updates $N$
		\STATE Initialize the online episode length $S$, the estimated model's  update frequency $\phi$, rollout batch size $R$ and length $H$
		\FOR{Epoch $t = 1 ... T$}
		\STATE Set the priority value of each transition $d\in D_{off}$ as $Pr(d,t)$
		\FOR{Step $\tau = 1 ... S$}
		\STATE Interact with the real environment for one step using $\pi_{on}$ to obtain the transition $d_{\tau} = (s_{\tau}, a_{\tau}, s_{\tau+1}, r_{\tau})$
		\STATE Compute the priority $Pr(d_\tau,t)$ and add the tuple $((s_{\tau}, a_{\tau}, s_{\tau+1}, r_{\tau}), Pr(d_\tau,t))$ to $\mathcal{D}_{on}$
		\IF{$\tau \mod \phi==0$} 
 		\STATE Train $\hat{M}$ until convergence on $\mathcal{D}_{off}\cup\mathcal{D}_{on}$ by prioritized sampling \cite{DBLP:journals/corr/SchaulQAS15}
	    \STATE Sample a batch of $R$ initial states $\{s_i\}_{i=1}^R$ from $\mathcal{D}_{off} \cup \mathcal{D}_{on}$ by prioritized sampling \cite{DBLP:journals/corr/SchaulQAS15}
		\STATE Generate $\{d_{i}\}_{i=1}^{R\times H}$ transitions by rollouting $\hat{M}$ using $\pi_{on}$ for $H$ steps starting from $\{s_i\}_{i=1}^R$
		\STATE Obtain the penalized rewards using equation (\ref{equ:uncertainty_reward}) (as done in MOPO) and add the updated transitions to $\mathcal{D}_{model}$
		\ENDIF
		\STATE Perform $N$ gradient updates on $\pi_{on}$ and $\hat{Q}_{on}$ using data uniformly sampled from $\mathcal{D}_{model}$
		\ENDFOR
		\ENDFOR
	\end{algorithmic}  
\end{algorithm*}

To boost data efficiency, we build our method upon the model-based offline RL framework. As aforementioned, distribution shift is the main concern in offline-to-online RL problems. A naive approach to this challenge is to directly transfer the policy, the value function, the environment dynamics model and the replay buffer in the offline stage to the online stage, and continue to fine-tune these functions using online interactions with the same offline algorithm. However, as we will see in Figure \ref{fig:ab1}, in some cases, the offline policy cannot be efficiently improved in the online stage.

To improve learning efficiency and control distribution shift, we design a prioritized sampling scheme that attaches different priorities to offline and online transitions at different online training epochs, and samples training batches using the prioritized experience replay approach \cite{DBLP:journals/corr/SchaulQAS15}. A general priority assignment scheme can be formulated as
\begin{equation}\label{equ:priority}
    Pr(d,t)=
    \begin{cases}
        f(d, t), &d\in \mathcal{D}_{off}\\
        1.0, &d\in \mathcal{D}_{on}
    \end{cases},
\end{equation}
where $d = (s,a,s',r)$ is a transition tuple in the dataset composed of both offline and online transitions, $t$ refers to the online training epoch, and $f(\cdot)$ is a priority function that maps its input $(d, t)$ into a value ranging from 0 to $\infty$. We denote $Pr(d, t)$ as priority here.

The offline-to-online RL performance is determined by the specific form of the function $f(d, t)$. In this paper, we adopt a priority function that only depends on the training epoch $t$, and its values linearly decays with $t$:
\begin{equation}
\label{prifunc}
    f(d, t)=\frac{1}{\alpha t},
\end{equation}
where $\alpha$ is a constant controlling the decaying speed of the offline data priority weight. Although this prioritized function is really simple, as we will see in section \ref{sec:experiment}, it can efficiently solve the major concerns of smooth transfer and fast adaption. Below we first intuitively discusses how the prioritized scheme encourages smooth transfer and fast adaptation, and then give a theoretical foundation in section \ref{subsec:theoretical_guarantee} to explain why we design MOORe in such a way.


\textbf{Smooth transfer.}
At the beginning of the online learning stage, on one hand, few online transitions are gathered; on the other hand, $Pr(d, e)\gg 0$, assigning large priority weights to offline data. As a result, offline data still occupies a dominant part. While in the offline training stage, the buffer only contains offline data, as online learning begins, the agent only suffers from a minor distribution shift.

\textbf{Fast adaptation.}
As online learning progresses, fast adaptation to the online stage becomes the main concern, as the performance of the current policy has surpassed that of the offline agent by a large margin and performance degradation in the early stage is no longer the primary contradiction. On one hand, the priority function (\ref{prifunc}) gives offline data a fairly small priority weight, resulting in a learning process almost relying on pure online data. On the other hand, the uncertainty term in Equation (\ref{equ:uncertainty_reward}) decays quickly in the online stage (discussed in subsection \ref{subsec:decay}) and the penalized reward function shift to a nearly regular one. Therefore, the online training process quickly turns into a MBPO-like training process, an efficient model-based online RL algorithm \cite{janner2019mbpo}.

Details of our proposed algorithm is summarized in Algorithm \ref{alg1}. The prioritized sampling scheme, carried out by the approach of prioritized experience replay \cite{DBLP:journals/corr/SchaulQAS15}, is used in both the model training process and the initial states sampling process.

\subsection{Theoretical Foundations}\label{subsec:theoretical_guarantee}
In this section, we provide an upper bound of the performance gap between two policies in two consecutive online learning epochs and discusses conditions and ways to reduce the gap.
The key idea is as follows: similar datasets would results in similar MDPs, and further similar performance of their optimal policies. We start by stating the closeness of optimal value functions of two similar MDPs $M_1$, $M_2$ in Theorem \ref{thm1}.

To measure the closeness of MDPs, we first introduce the $\ell_{1}$-norm distance. The $\ell_{1}$-norm distance between two probability distributions $p_{M_1}(\cdot|s,a)$ and $p_{M_2}(\cdot|s,a)$ is defined as
\begin{equation}
\begin{aligned}
    \delta_{\ell_{1}}\big(p_{M_1}(\cdot|s,a)&,p_{M_2}(\cdot|s,a)\big)\\
    &=\sum_{s'\in \mathcal{S}}|p_{M_1}(s'|s,a)-p_{M_2}(s'|s,a)|,
\end{aligned}
\end{equation}
where $p_{M_1},p_{M_2}$ represent the transition dynamics of MDPs $M_1,M_2$. We then define the distance between two transition dynamics $p_{M_1},p_{M_2}$, based on the $\ell_{1}$-norm distance between probabilities:
\begin{equation}
    D_{\ell_{1}}(p_{M_1},p_{M_2})=\max_{(s,a)\in \mathcal{S}\times\mathcal{A}}\delta_{\ell_1}\big(p_{M_1}(\cdot|s,a),p_{M_2}(\cdot|s,a)\big).
\end{equation}
In finite-horizon MDPs, define $V_{M,h}^*(s)$ as the value function on state $s$ with the optimal policy $\pi^*$ at horizon $h$. Define the maximum reward $r_{max}=\max_{(s,a)\in\mathcal{S}\times\mathcal{A}}|r(s,a)|$ and the maximum value $V_{max}=\max_{s\in\mathcal{S},\pi\in \Pi}V^\pi(s)$ ($\Pi$ is the policy space). For simplicity, we regard $r_{max},V_{max}$ as the upper bound of $\hat{M}$ and $\tilde{M}$.
\begin{theorem}
\label{thm1}
Let $M_1,M_2$ be two finite-horizon MDPs with the same reward function $r$. Then the distance of their optimal value function $V^*_{M_1,h}(s)$ and $V^*_{M_2,h}(s)$ is bounded by
\begin{equation}\label{equ:them1}
    |V^*_{M_1,h}(s)-V^*_{M_2,h}(s)|\le\epsilon_h,
\end{equation}
 where $\epsilon_h=D_{\ell_{1}}(p_{M_1},p_{M_2})(r_{max}+V_{max})(H-h),\ \forall h\in[H],s\in\mathcal{S}$.
\end{theorem}
The proof can be found in Appendix \ref{proofs}.

Theorem \ref{thm1} conveys the fact that, when two MDPs are close enough in terms of the $\ell_{1}$ distance of transition dynamics and equivalence of reward function, the resulting optimal value functions would be bounded. The assumptions always hold because we consider two successive epochs, where the buffer and relevant networks rarely changes.

In the second step, we use this closeness of the optimal value functions to prove the closeness of expected discounted return between policies generated by two close MDPs.

By combining Equation (\ref{equ:target}) and Equation (\ref{equ:them1}), we have
\begin{equation}\label{equ:eta}
\begin{aligned}
|\eta_{M_1}(\pi_{M_1}^*)&-\eta_{M_2}(\pi_{M_2}^*)|\\
&\le D_{\ell_{1}}(p_{M_1},p_{M_2})(r_{max}+V_{max})H.
\end{aligned}
\end{equation}
In our offline-to-online setting, we specifically considers the uncertainty-penalized estimated MDP $\tilde{M}=(\mathcal{S},\mathcal{A},\hat{p},\tilde{r},\mu_0,\gamma)$.  Let $M_1=\tilde{M}_{t},M_2=\tilde{M}_{t+1}$ in Equation (\ref{equ:eta}), where $t,t+1$ refer to two consecutive epochs in the offline-to-online training stage, we have
\begin{equation}
\begin{aligned}
|\eta_{\tilde{M}_{t}}(\pi_{\tilde{M}_{t}}^*)&-\eta_{\tilde{M}_{t+1}}(\pi_{\tilde{M}_{t+1}}^*)| \\
&\le D_{\ell_{1}}(p_{\tilde{M}_{t}},p_{\tilde{M}_{t+1}})(r_{max}+V_{max})H.
\end{aligned}
\end{equation}

For the following proofs, we introduce $\rho_{M}^\pi(s,a)$, which is defined as the discounted occupancy measure of policy $\pi$ under the dynamics of MDP $M$.

\begin{lemma}
\label{lemma1}
In the uncertainty-penalized MDP, define $U_{\hat{M}}(\pi)=\overline{\mathbb{E}}_{(s,a)\sim\rho^\pi_{\hat{M}}}[u_{\hat{M}}(s,a)]$, 
which characterizes how erroneous the model is along trajectories induced by $\pi$. Then 
\begin{equation}
    \begin{aligned}
    |\eta_{\hat{M}_t}(\pi_{\tilde{M}_t}^*)&-\eta_{\hat{M}_{t+1}}(\pi_{\tilde{M}_{t+1}}^*)|\\
    &\le D_{\ell_{1}}(p_{\hat{M}_{t}},p_{\hat{M}_{t+1}})(r_{max}+V_{max})H\\
    &+\lambda|U_{\hat{M}_t}(\pi_{\tilde{M}_t}^*)-U_{\hat{M}_{t+1}}(\pi_{\tilde{M}_{t+1}}^*)|.
\end{aligned}
\end{equation}
\end{lemma}

Next, we introduce the Telescoping lemma mentioned in the paper of MOPO, which gives the closeness proof between the performance of a policy under different dynamics.

\begin{lemma}
\label{tele}
(Telescoping lemma) \cite{yu2020mopo,DBLP:journals/corr/abs-1807-03858}. Let $M$ and $\hat{M}$ be two MDPs with the same reward $r(s,a)$, but different dynamics $p_M$ and $p_{\hat{M}}$ respectively. Let
\begin{equation}
\begin{aligned}
 G_{\hat{M}}&^\pi(s,a):=\\
 &\mathbb{E}_{s'\sim p_{\hat{M}}(s,a)}[V^\pi_M(s')]-\mathbb{E}_{s'\sim p_M(s,a)}[V^\pi_M(s')],
\end{aligned}
\end{equation}
Then,
\begin{equation}
    \eta_{\hat{M}}(\pi)-\eta_M(\pi)=\gamma\mathbb{E}_{(s,a)\sim\rho_{\hat{M}}^\pi}[G_{\hat{M}}^\pi(s,a)].
\end{equation}
\end{lemma}
For each $s\in\mathcal{S},a\in\mathcal{A}$, a $\ell_{1}$-based bound of $|G_{\hat{M}}^\pi(s,a)|$ is
\begin{equation}
\label{Gineq}
    |G_{\hat{M}}^\pi(s,a)|\le \frac{1}{2}V_{max}\delta_{\ell_{1}}(p_{\hat{M}}(s,a),p_M(s,a)).
\end{equation}

Combining the bound between estimated MDP and real MDP with Theorem \ref{thm1}, we are able to prove Theorem \ref{thm2}.
\begin{theorem}
\label{thm2}
Suppose $M$ is the real MDP. $\hat{M}_t,\hat{M}_{t+1}$ are two estimated MDPs, and $\tilde{M}_t,\tilde{M}_{t+1}$ are their uncertainty-penalized MDPs. $\pi_{\tilde{M}_1}^*$, $\pi_{\tilde{M}_2}^*$ are the optimal policies under $\tilde{M}_1$, $\tilde{M}_2$, respectively. Then the difference between their expected discounted returns under MDP $M$ is bounded by
\begin{equation}
    \begin{aligned}
    |\eta&_{M}(\pi_{\tilde{M}_t}^*)-\eta_{M}(\pi_{\tilde{M}_{t+1}}^*)|\\
    &\le D_{\ell_{1}}(p_{\hat{M}_t},p_{\hat{M}_{t+1}})(r_{max}+V_{max})H\\
    &+\Big[\mathbb{E}_{(s,a)\sim\rho_{\hat{M}_t}^{\pi_{\tilde{M}_t}^*}(s,a)}\delta_{\ell_{1}}\big(p_{\hat{M}_t}(s,a),p_M(s,a)\big)\\
    &+\mathbb{E}_{(s,a)\sim\rho_{\hat{M}_{t+1}}^{\pi_{\tilde{M}_{t+1}}^*}(s,a)}\delta_{\ell_{1}}\big(p_{\hat{M}_{t+1}}(s,a),p_M(s,a)\big)\Big]\frac{1}{2}\gamma V_{max}\\
    &+\lambda|U_{\hat{M}_t}(\pi_{\tilde{M}_t}^*)-U_{\hat{M}_{t+1}}(\pi_{\tilde{M}_{t+1}}^*)|.\\
\end{aligned}
\end{equation}
\end{theorem}
The proof can also be found in Appendix \ref{proofs}.

\begin{remark}\label{rem:remark}
According to Theorem \ref{thm2}, the performance gap between two optimal policies in two consecutive epochs is narrow if
\begin{enumerate}[label=(\roman*)]
    \item $\hat{M}_t$ and $\hat{M}_{t+1}$ are close to each other in terms of $D_{\ell_1}(p_{\hat{M}_t},p_{\hat{M}_{t+1}})$;
    \item $\hat{M}_t$ and $\hat{M}_{t+1}$ are close to the real MDP $M$ in \textbf{observed state-action pairs} having distributions $\rho_{p_{M_1}}^{\pi_{M_1}^*}(s,a)$ and $\rho_{p_{M_2}}^{\pi_{M_2}^*}(s,a)$,  respectively;
    \item $U_{\hat{M}_t}(\pi_{\tilde{M}_t}^*)$ and $U_{\hat{M}_{t+1}}(\pi_{\tilde{M}_{t+1}}^*)$ are close to each other. It describes the extent of the model error along trajectories induced by $\pi_{\tilde{M}_t}^*$ and $\pi_{\tilde{M}_{t+1}}^*$.
\end{enumerate}

(\romannumeral1) is easy to meet with as long as the two datasets used to train $\hat{M}_t$ and $\hat{M}_{t+1}$ are similar. Since empirically, Equation (\ref{prifunc}) ensures that the datasets used to train dynamics models between two consecutive steps are almost the same as each other.
(\romannumeral2) naturally holds because the model estimation is relatively accurate for observed state-action pairs.
For condition (\romannumeral3), although we can assume that the estimated uncertainty $u_{\hat{M}_t}(s,a),u_{\hat{M}_{t+1}}(s,a)$ is similar due to the similarity between $\hat{M}_t$ and $\hat{M}_{t+1}$, we cannot reach the conclusion that $|U_{\hat{M}_t}(\pi_{\tilde{M}_t}^*)-U_{\hat{M}_{t+1}}(\pi_{\tilde{M}_{t+1}}^*)|$ is also small, since the similarity between the occupancy measures $\rho_{\hat{M}_t}^{\pi_{\tilde{M}_t}^*}$ and $\rho_{\hat{M}_{t+1}}^{\pi_{\tilde{M}_{t+1}}^*}$ cannot be ensured without further assumptions (A corner case is that a given MDP may have multiple optimal policies and their  occupancy measures may be different while the performance is the same). However, in most cases, similar MDPs can yield similar occupancy measures. So here we make an assumption that the uncertainty difference is small in real environments, which will be empirically verified in Section \ref{Supplementary}.
\end{remark}

\begin{figure*}[htbp]
    \centering
    \includegraphics[width=\textwidth]{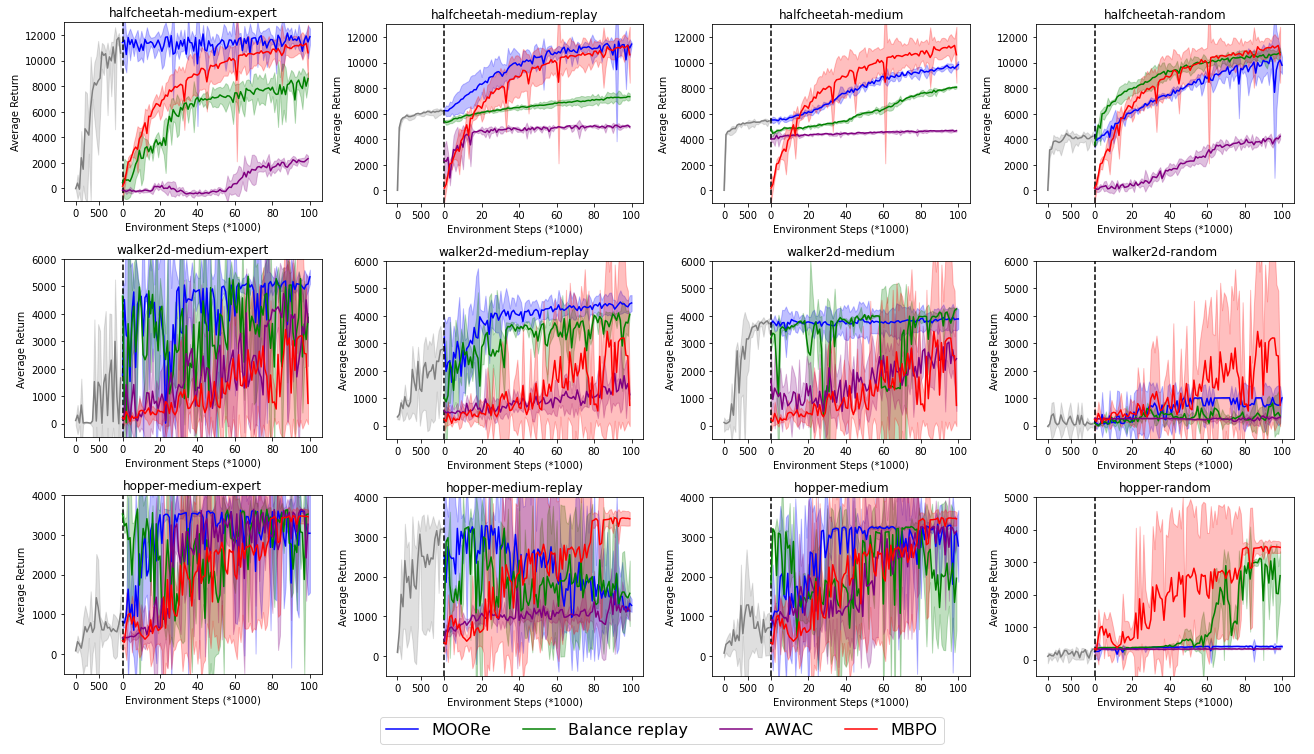}
    \caption{Results on MuJoCo locomotion tasks. The gray curves on the left side of dotted lines are the offline learning curves of MOORe, and those on the right-hand side are online learning curves of different algorithms.}
    \label{fig1}
\end{figure*}

In summary, as long as the learned models between two consecutive steps are close to each other, the corresponding policies' performance gap can be small. Our proposed prioritized sampling scheme ensures that changes in the datasets are minor and thus ensures transfer smoothness.


\subsection{Influence of Conservatism}
\label{subsec:decay}
The offline RL algorithm trains in a conservative manner, but online RL does not need to be conservative. Directly abandoning conservatism when the agent enters the online stage may harm transfer smoothness, while sticking to conservatism may harm efficiency.

In this paper, we continue to use the penalized reward function as defined in Equation \ref{equ:uncertainty_reward} even in the online training stage. As we will see in Figure \ref{fig:ab2} in Section \ref{sec:experiment}, MOORe finally acts like an online RL algorithm, because the uncertainty term decreases quickly as the online training progress continues and thus has minor effect on the final performance.

\section{Experiments}
\label{sec:experiment}
Below we first present the results of MOORe in twelve D4RL benchmarks in the MuJoCo control tasks \cite{DBLP:journals/corr/abs-2004-07219}, which are widely used in both offline and offline-to-online settings. Then we conduct ablation studies to investigate how critical components of MOORe affect the final performance. 

We follow the same hyper-parameter configurations of MOPO in the offline learning stage. In the online learning stage, the maximum online training epoch is $T=100$, and in each epoch, $S=1000$ steps of online transitions are collected. The model training frequency is set to $\phi=250$. The number of model rollouts is $R=100000$. The number of gradient updates is set to $N=20$, which is large enough for convergence. The above hyper-parameter configurations in the online setting are adapted from MBPO. The rollout length $H$ and penalty coefficient $\lambda$ (defined in Section \ref{mboorl}) follow the default setting in MOPO (varying for different tasks).

\subsection{Main Results}\label{subsec:main_results}
We compare our algorithm with three baselines, where balance replay \cite{lee2021offlinetoonline} and AWAC \cite{nair2021awac} are existing offline-to-online algorithms, and MBPO \cite{janner2019mbpo} is a pure online RL algorithm.

As illustrated in Figure \ref{fig1} (we only plot the offline learning curves of our proposed method, for clarity), MOORe outperforms previous offline-to-online methods in most cases. On one hand, it converges much faster: previous best-performed methods requires around 250 epochs until converge, while ours only need 100 epochs at most. In many cases, less than 50 epochs is enough convergence. On the other hand, the learning curves of MOORe are smooth during the offline to the online transferring procedure, demonstrating its ability to robustly switch to the online training stage without performance degradation, unlike some counterexamples of baseline algorithms (e.g., balance replay algorithm in the \emph{hopper-medium-replay} dataset). Consequently, we can draw the conclusion that our proposed algorithm MOORe is capable of solving both challenges of fast adaption and smooth transfer in the offline-to-online learning setting. 

Failure cases of MOORe can be found on the \emph{random} benchmarks (\emph{walker2d-random} and \emph{hopper-random}), which are generated by random policies \cite{DBLP:journals/corr/abs-2004-07219}. We guess that contributes to the collapsed policy in the offline learning stage. A ruined policy can not be improved by MOORe. Fortunately, random datasets can be rarely encountered in real applications as the policy used to collect data should at least be capable of finishing some simple tasks.

\subsection{Ablation study}\label{subsec:ablation_study}
Below we investigate how different components of the prioritized sampling scheme affect the performance of MOORe.

\paragraph{Different Prioritized Sampling Schemes.}
The prioritized sampling scheme plays a critical role in the success of MOORe. To verify this, we compare it with three other commonly used sampling schemes. The first sampling scheme is \emph{uniform sampling}, meaning that all of offline and online transitions are sampled by a equal probability. \emph{Half-half sampling} means that in each training batch, half of the samples are from offline transitions and half online. \emph{Pure online sampling} only uses the online interactions in the online learning stage.

\begin{figure}
    \centering
    \includegraphics[width=0.48\textwidth]{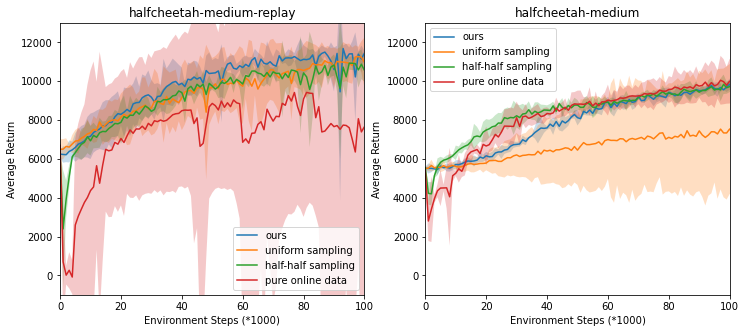}
    \caption{Different data sampling schemes.}
    \label{fig:ab1}
\end{figure}

As shown in Figure \ref{fig:ab1}, the key weakness of the half-half sampling and the pure online sampling schemes is the sudden performance drop at the beginning of the online stage. But their overall speed of convergence is fast. In contrast, the uniform sampling scheme yields smooth learning curves at the beginning, but fails to adapt to online data efficiently.

Our proposed sampling scheme pays attention to transfer smoothness in the early online learning stage and fast adaptation in the late stage. In the first few online learning epochs, the policy improves itself while maintaining the majority of offline knowledge. When it totally adapts to the online environment, it can also converge with a relatively high speed. Although it is slower than the half-half sampling and the pure online sampling schemes, the gap is narrow and acceptable.

\begin{figure}
    \centering
    \includegraphics[width=0.48\textwidth]{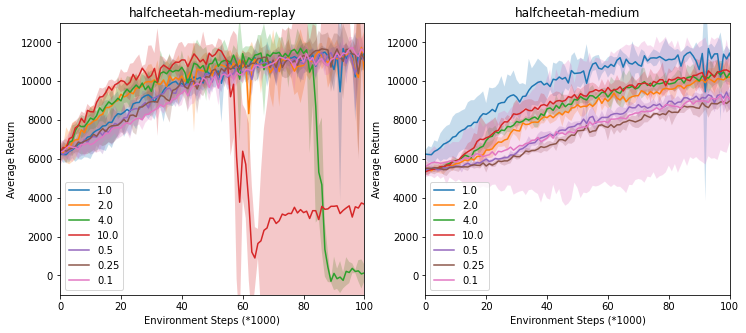}
    \caption{Learning curves under different decay rate ($\alpha$).}
    \label{fig:ab3}
\end{figure}

\begin{figure}
    \centering
    \includegraphics[width=0.48\textwidth]{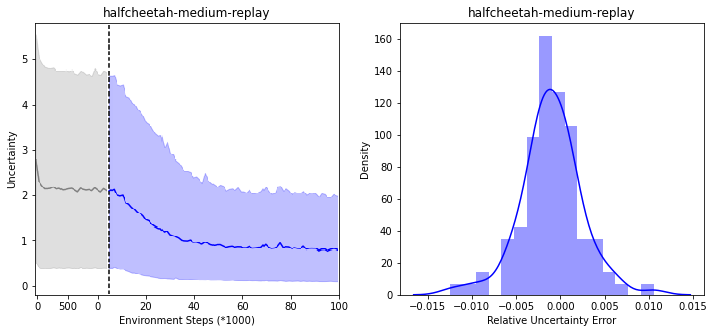}
    \caption{The LHS figure shows how the uncertainty (defined in MOPO) changes with training steps. The shadowed area is the confidence interval between 20\% and 80\%. The RHS figure plots the distribution of Relative Uncertainty Error. The x-axis is the value and the y-axis is the probability density function.}
    \label{fig:ab2}
\end{figure}

\paragraph{Robustness to Priority Hyper-parameters.}
We also investigate the effect of the hyper-parameter $\alpha$ in Equation (\ref{equ:priority}). Specifically, we set $\alpha$ to $[0.1,0.25,0.5,1.0, 2.0,4.0,10.0]$ and re-implement MOORe on part of the \emph{Halfcheetah-medium-replay} benchmarks. As shown in Figure \ref{fig:ab2}, MOORe keeps the state-of-the-art performance under a wide range of hyper-parameter configurations and collapses only under several extreme cases. Therefore, we can draw the conclusion that our proposed simple prioritized sampling scheme is robust under a spectrum of hyper-parameter configurations.

\subsection{Effect of Penalty Coefficient}
\label{Supplementary}
As mentioned before, the conservatism term in MOPO may cause side effect on MOORe in the online learning phase. Figure \ref{fig:ab3} demonstrates the curve of the empirical expectation of the uncertainty term in Equation (\ref{equ:uncertainty_reward}) during the entire online learning stage. As can be seen, its fast decay suggests that the conservatism term brings little influence on the online policy, especially in the late learning stage.

We also compute the Relative Uncertainty Error, defined as $|U_{\hat{M}_t}-U_{\hat{M}_{t+1}}(\pi_{\tilde{M}_t}^*)|/\eta_{\hat{M}_t}(\pi_{\tilde{M}_t}^*)$, to verify the assumption made in the remark of Theorem \ref{thm2}. As shown in Figure \ref{fig:ab3}, the difference is indeed small when compared to the expected return, which is in accordance with our assumption that the difference of the expected uncertainty values between two consecutive epochs is small and thus the transfer smoothness can be guaranteed as long as the estimated models between two consecutive epochs are close to each other.

\section{Conclusion}\label{sec:conclusion}
In this paper, we look into the problem of offline-to-online RL, where agents are first trained until convergence on the offline dataset, and then fine-tuned online. We start by noticing the importance of smooth transfer and fast adaption from offline to online stages. To tackle these challenges, we introduced a Model-based Offline-to-Online RL algorithm (MOORe), which features a prioritized data sampling scheme in the model-based framework to provide high data efficiency in online the training stage as well as ensuring transfer smoothness with theoretical foundations. Experiment results show that our algorithm empowers smooth transfer of the offline policy to the online setting and continues to improve its performance efficiently, achieving state-of-the-art performance on the D4RL benchmarks from the MuJoCo Control tasks. 


\newpage

\bibliographystyle{named}
\bibliography{ijcai19}

\begin{thebibliography}{}

\bibitem[\protect\citeauthoryear{Anonymous}{2022}]{anonymous2022adaptive}
Anonymous.
\newblock Adaptive behavior cloning regularization for stable offline-to-online
  reinforcement learning.
\newblock In {\em Submitted to The Tenth International Conference on Learning
  Representations}, 2022.
\newblock under review.

\bibitem[\protect\citeauthoryear{Fu \bgroup \em et al.\egroup
  }{2020}]{DBLP:journals/corr/abs-2004-07219}
Justin Fu, Aviral Kumar, Ofir Nachum, George Tucker, and Sergey Levine.
\newblock {D4RL:} datasets for deep data-driven reinforcement learning.
\newblock {\em CoRR}, abs/2004.07219, 2020.

\bibitem[\protect\citeauthoryear{Fujimoto \bgroup \em et al.\egroup
  }{2018}]{fujimoto2018offpolicy}
Scott Fujimoto, David Meger, and Doina Precup.
\newblock Off-policy deep reinforcement learning without exploration, 2018.

\bibitem[\protect\citeauthoryear{Gottesman \bgroup \em et al.\egroup
  }{2019}]{healthcare}
Omer Gottesman, Fredrik Johansson, Matthieu Komorowski, Aldo Faisal, Finale
  Doshi-Velez David~Sontag, , and Leo~Anthony Celi.
\newblock Guidelines for reinforcement learning in healthcare.
\newblock In {\em Nature medicine}, page 25(1):16–18, 2019.

\bibitem[\protect\citeauthoryear{Hester \bgroup \em et al.\egroup
  }{2017}]{hester2017deep}
Todd Hester, Matej Vecerik, Olivier Pietquin, Marc Lanctot, Tom Schaul, Bilal
  Piot, Dan Horgan, John Quan, Andrew Sendonaris, Gabriel Dulac-Arnold, Ian
  Osband, John Agapiou, Joel~Z. Leibo, and Audrunas Gruslys.
\newblock Deep q-learning from demonstrations, 2017.

\bibitem[\protect\citeauthoryear{Janner \bgroup \em et al.\egroup
  }{2019}]{janner2019mbpo}
Michael Janner, Justin Fu, Marvin Zhang, and Sergey Levine.
\newblock When to trust your model: Model-based policy optimization.
\newblock In {\em Advances in Neural Information Processing Systems}, 2019.

\bibitem[\protect\citeauthoryear{Kang \bgroup \em et al.\egroup
  }{2018}]{pmlr-v80-kang18a}
Bingyi Kang, Zequn Jie, and Jiashi Feng.
\newblock Policy optimization with demonstrations.
\newblock In Jennifer Dy and Andreas Krause, editors, {\em Proceedings of the
  35th International Conference on Machine Learning}, volume~80 of {\em
  Proceedings of Machine Learning Research}, pages 2469--2478. PMLR, 10--15 Jul
  2018.

\bibitem[\protect\citeauthoryear{Kidambi \bgroup \em et al.\egroup
  }{2020}]{kidambi2020morel}
Rahul Kidambi, Aravind Rajeswaran, Praneeth Netrapalli, and Thorsten Joachims.
\newblock Morel : Model-based offline reinforcement learning, 2020.

\bibitem[\protect\citeauthoryear{Kumar \bgroup \em et al.\egroup
  }{2019}]{kumar2019stabilizing}
Aviral Kumar, Justin Fu, George Tucker, and Sergey Levine.
\newblock Stabilizing off-policy q-learning via bootstrapping error reduction,
  2019.

\bibitem[\protect\citeauthoryear{Kumar \bgroup \em et al.\egroup
  }{2020}]{kumar2020conservative}
Aviral Kumar, Aurick Zhou, George Tucker, and Sergey Levine.
\newblock Conservative q-learning for offline reinforcement learning, 2020.

\bibitem[\protect\citeauthoryear{Lee \bgroup \em et al.\egroup
  }{2021}]{lee2021offlinetoonline}
Seunghyun Lee, Younggyo Seo, Kimin Lee, Pieter Abbeel, and Jinwoo Shin.
\newblock Offline-to-online reinforcement learning via balanced replay and
  pessimistic q-ensemble, 2021.

\bibitem[\protect\citeauthoryear{Liu \bgroup \em et al.\egroup
  }{2020}]{DBLP:journals/corr/abs-2007-08202}
Yao Liu, Adith Swaminathan, Alekh Agarwal, and Emma Brunskill.
\newblock Provably good batch reinforcement learning without great exploration.
\newblock {\em CoRR}, abs/2007.08202, 2020.

\bibitem[\protect\citeauthoryear{Nair \bgroup \em et al.\egroup
  }{2021}]{nair2021awac}
Ashvin Nair, Abhishek Gupta, Murtaza Dalal, and Sergey Levine.
\newblock Awac: Accelerating online reinforcement learning with offline
  datasets, 2021.

\bibitem[\protect\citeauthoryear{Reddy \bgroup \em et al.\egroup
  }{2019}]{reddy2019sqil}
Siddharth Reddy, Anca~D. Dragan, and Sergey Levine.
\newblock Sqil: Imitation learning via reinforcement learning with sparse
  rewards, 2019.

\bibitem[\protect\citeauthoryear{Schaul \bgroup \em et al.\egroup
  }{2016}]{DBLP:journals/corr/SchaulQAS15}
Tom Schaul, John Quan, Ioannis Antonoglou, and David Silver.
\newblock Prioritized experience replay.
\newblock In Yoshua Bengio and Yann LeCun, editors, {\em 4th International
  Conference on Learning Representations, {ICLR} 2016, San Juan, Puerto Rico,
  May 2-4, 2016, Conference Track Proceedings}, 2016.

\bibitem[\protect\citeauthoryear{Wu \bgroup \em et al.\egroup
  }{2019}]{wu2019behavior}
Yifan Wu, George Tucker, and Ofir Nachum.
\newblock Behavior regularized offline reinforcement learning, 2019.

\bibitem[\protect\citeauthoryear{Xu \bgroup \em et al.\egroup
  }{2018}]{DBLP:journals/corr/abs-1807-03858}
Huazhe Xu, Yuanzhi Li, Yuandong Tian, Trevor Darrell, and Tengyu Ma.
\newblock Algorithmic framework for model-based reinforcement learning with
  theoretical guarantees.
\newblock {\em CoRR}, abs/1807.03858, 2018.

\bibitem[\protect\citeauthoryear{Yu \bgroup \em et al.\egroup
  }{2018}]{DBLP:journals/corr/abs-1805-04687}
Fisher Yu, Wenqi Xian, Yingying Chen, Fangchen Liu, Mike Liao, Vashisht
  Madhavan, and Trevor Darrell.
\newblock {BDD100K:} {A} diverse driving video database with scalable
  annotation tooling.
\newblock {\em CoRR}, abs/1805.04687, 2018.

\bibitem[\protect\citeauthoryear{Yu \bgroup \em et al.\egroup
  }{2020}]{yu2020mopo}
Tianhe Yu, Garrett Thomas, Lantao Yu, Stefano Ermon, James Zou, Sergey Levine,
  Chelsea Finn, and Tengyu Ma.
\newblock Mopo: Model-based offline policy optimization, 2020.

\bibitem[\protect\citeauthoryear{Yu \bgroup \em et al.\egroup
  }{2021}]{yu2021combo}
Tianhe Yu, Aviral Kumar, Rafael Rafailov, Aravind Rajeswaran, Sergey Levine,
  and Chelsea Finn.
\newblock Combo: Conservative offline model-based policy optimization, 2021.

\end{thebibliography}

\newpage

\appendix

\section{Theorem proofs}
\label{proofs}
Below is the full proof of Theorem \ref{thm1}. For ease of discussion, we only consider finite-horizon MDPs.
\begin{proof}
We begin our proof from the last horizon $h=H$, and then use the closeness under horizon $h$ to prove the closeness under horizon $h-1$.

For the last horizon $h=H$, $V_{M,H}^{*}(s)=0\ \forall M$ because it is the terminal state, so $||V^*_{M_1,H}(s)-V^*_{M_2,H}||_\infty\le 0=\epsilon_H$. Suppose
\begin{equation}
    \forall s\in\mathcal{S}_h,||V^*_{M_1,h}(s)-V^*_{M_2,h}||_\infty\le\epsilon_h.
\end{equation}
We need to prove
\begin{equation}
    \forall s\in\mathcal{S}_{h-1},||V^*_{M_1,h-1}(s)-V^*_{M_2,h-1}||_\infty\le\epsilon_{h-1}.
\end{equation}
It is equivalent to prove
\begin{equation}\label{equ:left_right}
    -\epsilon_{h-1}\le V^*_{M_2,h-1}(s)-V^*_{M_1,h-1}(s)\le\epsilon_{h-1}.
\end{equation}
For simplicity but without loss of generality, we prove the right part inequality in the above equation.
\begin{equation}
    \begin{aligned}
    LHS&=\max_{a\in\mathcal{A}}\{\sum_{s'\in\mathcal{S}_h}p_{M_2}(s'|s,a)(r(s,a)+\gamma V^*_{M_2,h}(s'))\}\\
    &-\max_{a\in\mathcal{A}}\{\sum_{s'\in\mathcal{S}_h}p_{M_1}(s'|s,a)(r(s,a)+\gamma V^*_{M_1,h}(s'))\}\\
    =&\max_{a\in\mathcal{A}}\{\sum_{s'\in\mathcal{S}_h}[p_{M_1}(s'|s,a)(r(s,a)+\gamma V^*_{M_1,h}(s'))\\
    &+(p_{M_2}(s'|s,a)-p_{M_1}(s'|s,a))r(s,a)\\
    &+p_{M_2}(s'|s,a)V^*_{M_2,h}(s')-p_{M_1}(s'|s,a)V^*_{M_1,h}(s')]\}\\
    &-\max_{a\in\mathcal{A}}\{\sum_{s'\in\mathcal{S}_h}p_{M_1}(s'|s,a)(r(s,a)+\gamma V^*_{M_1,h}(s'))\}\\
    \le&\max_{a\in\mathcal{A}}\{\sum_{s'\in\mathcal{S}_h}p_{M_1}(s'|s,a)(r(s,a)+\gamma V^*_{M_1,h}(s'))\}\\
    &+\max_{a\in\mathcal{A}}\{\sum_{s'\in\mathcal{S}_h}[(p_{M_2}(s'|s,a)-p_{M_1}(s'|s,a))r(s,a)\\
    &+p_{M_2}(s'|s,a)V^*_{M_2,h}(s')-p_{M_1}(s'|s,a)V^*_{M_1,h}(s')]\}\\
    &-\max_{a\in\mathcal{A}}\{\sum_{s'\in\mathcal{S}_h}p_{M_1}(s'|s,a)(r(s,a)+\gamma V^*_{M_1,h}(s'))\}\\
    =&\max_{a\in\mathcal{A}}\{\sum_{s'\in\mathcal{S}_h}[(p_{M_2}(s'|s,a)-p_{M_1}(s'|s,a))r(s,a)\\
    &+p_{M_2}(s'|s,a)V^*_{M_2,h}(s')-p_{M_1}(s'|s,a)V^*_{M_1,h}(s')]\}\\
    \le&\max_{a\in\mathcal{A}}\{\sum_{s'\in\mathcal{S}_h}|p_{M_2}(s'|s,a)-p_{M_1}(s'|s,a)|\}r_{max}\\
    &+\max_{a\in\mathcal{A}}\{\sum_{s'\in\mathcal{S}_h}(p_{M_2}(s'|s,a)-p_{M_1}(s'|s,a))V^*_{M_2,h}(s')\\
    &+p_{M_1}(s'|s,a)(V^*_{M_2,h}(s')-V^*_{M_1,h}(s'))\}\\
    \le&\max_{a\in\mathcal{A}}\{D_{\ell_{1}}(p_{M_1}(\cdot|s,a),p_{M_2}(\cdot|s,a))\}(r_{max}+V_{max})\\
    &+1\cdot(V^*_{M_2,h}(s')-V^*_{M_1,h}(s'))\}\\
    \le&D_{\ell_{1}}(p_{M_1},p_{M_2})(r_{max}+V_{max})+1\cdot\epsilon(h)\}\\
    =&D_{\ell_{1}}(p_{M_1},p_{M_2})(r_{max}+V_{max})\\
    &+D_{\ell_{1}}(p_{M_1},p_{M_2})(r_{max}+V_{max})(H-h)\\
    =&D_{\ell_{1}}(p_{M_1},p_{M_2})(r_{max}+V_{max})(H-h+1).
\end{aligned}
\end{equation}

\end{proof}

The proof of Lemma \ref{lemma1} is as follows.
\begin{proof}
    First prove that $\forall$ estimated MDP $\hat{M}$ and its relevant uncertainty penalized MDP $\tilde{M}$,
    \begin{equation}
        \eta_{\tilde{M}}(\pi_{\tilde{M}}^*)=\eta_{\hat{M}}(\pi_{\tilde{M}}^*)-\lambda U_{\hat{M}}(\pi_{\tilde{M}}^*).
    \end{equation}
    We know that $\tilde{M}$ and $\hat{M}$ shares the same transition dynamics $p$, but different reward functions $\tilde{r}(s,a)=\hat{r}(s,a)-\lambda u(s,a)$. Therefore,
    \begin{equation}
        \begin{aligned}
        \eta_{\tilde{M}}(\pi_{\tilde{M}}^*)=&\mathbb{E}_{(s,a)\sim\rho_{{\tilde{M}}}^{\pi_{\tilde{M}}^*}(s,a)}[\tilde{r}(s,a)]\\
        =&\mathbb{E}_{(s,a)\sim\rho_{{\hat{M}}}^{\pi_{\tilde{M}}^*}(s,a)}[\hat{r}(s,a)-\lambda u(s,a)]\\
        =&\mathbb{E}_{(s,a)\sim\rho_{{\hat{M}}}^{\pi_{\tilde{M}}^*}(s,a)}\hat{r}(s,a)-\lambda\mathbb{E}_{(s,a)\sim\rho_{{\hat{M}}}^{\pi_{\tilde{M}}^*}(s,a)}u(s,a)\\
        =&\eta_{\hat{M}}(\pi_{\tilde{M}}^*)-\lambda U_{\hat{M}}(\pi_{\tilde{M}}^*).\\
    \end{aligned}
    \end{equation}
    Next we'll decompose $ |\eta_{\hat{M}_t}(\pi_{\tilde{M}_t}^*)-\eta_{\hat{M}_{t+1}}(\pi_{\tilde{M}_{t+1}}^*)|$ to get the final bound.
    \begin{equation}
        \begin{aligned}
         |\eta_{\hat{M}_t}&(\pi_{\tilde{M}_t}^*)-\eta_{\hat{M}_{t+1}}(\pi_{\tilde{M}_{t+1}}^*)|\\
         =&|(\eta_{\tilde{M}_t}(\pi_{\tilde{M}_t}^*)+\lambda U_{\hat{M}_t}(\pi_{\tilde{M}_t}^*))\\
         &\qquad-(\eta_{\tilde{M}_{t+1}}(\pi_{\tilde{M}_{t+1}}^*)+\lambda U_{\hat{M}_{t+1}}(\pi_{\tilde{M}_{t+1}}^*))|\\
         \le&|\eta_{\tilde{M}_t}(\pi_{\tilde{M}_t}^*)-\eta_{\tilde{M}_{t+1}}(\pi_{\tilde{M}_t}^*)|\\
         &\qquad+\lambda|U_{\hat{M}_t}(\pi_{\tilde{M}_t}^*)-U_{\hat{M}_{t+1}}(\pi_{\tilde{M}_{t+1}}^*)|\\
         \le& D_{\ell_{1}}(p_{\tilde{M}_{t}},p_{\tilde{M}_{t+1}})(r_{max}+V_{max})H\\
         &\qquad+\lambda|U_{\hat{M}_t}(\pi_{\tilde{M}_t}^*)-U_{\hat{M}_{t+1}}(\pi_{\tilde{M}_{t+1}}^*)|\\
         =& D_{\ell_{1}}(p_{\hat{M}_{t}},p_{\hat{M}_{t+1}})(r_{max}+V_{max})H\\
         &\qquad+\lambda|U_{\hat{M}_t}(\pi_{\tilde{M}_t}^*)-U_{\hat{M}_{t+1}}(\pi_{\tilde{M}_{t+1}}^*)|.\\
    \end{aligned}
    \end{equation}
    
\end{proof}
\paragraph{Remarks on Equation (\ref{Gineq})}
The original inequality in Equation (\ref{Gineq}) is introduced in MOPO \cite{yu2020mopo}, using the Total Variation Distance as the distance measure, instead of $\delta_{\ell_1}(\cdot,\cdot)$. The Total Variation Distance is defined as $\delta_{TV}(P,Q)=\sup_{A\in\mathcal{F}}|P(A)-Q(A)|$ where $P,Q$ are two probability measures on a sigma algebra $\mathcal{F}$. It is written as 
\begin{equation}
    |G_{\hat{M}}^\pi(s,a)|\le V_{max}\delta_{TV}(p_{\hat{M}}(s,a),p_M(s,a)).
\end{equation}
And the version in Equation (\ref{Gineq}) has the same meaning as $\delta_{TV}(P,Q)=\frac{1}{2}\delta_{\ell_1}(P,Q)$, because $\delta_{TV}(P,Q)=\frac{1}{2}\delta_{\ell_1}(P,Q)$ when the sigma-algebra $\mathcal{F}$ is countable.

The proof of Theorem \ref{thm2} is carried out according to Lemma \ref{lemma1} and Lemma \ref{tele}.
\begin{proof}
    The first step considers $\eta_{\hat{M}_t}(\pi_{\tilde{M}_t}^*),\eta_{\hat{M}_{t+1}}(\pi_{\tilde{M}_{t+1}}^*)$ to build the bound. And in the second step, Lemma \ref{tele} is used to bound $|\eta_{M}(\pi_{\tilde{M}_t}^*)-\eta_{\hat{M}_t}(\pi_{\tilde{M}_t}^*)|$ and $|\eta_{M}(\pi_{\tilde{M}_{t+1}}^*)-\eta_{\hat{M}_{t+1}}(\pi_{\tilde{M}_{t+1}}^*)|$, while $|\eta_{\hat{M}_t}(\pi_{\tilde{M}_t}^*)-\eta_{\hat{M}_{t+1}}(\pi_{\tilde{M}_{t+1}}^*)|$ is bounded by Lemma \ref{lemma1}.
\begin{equation}
    \begin{aligned}
    |&\eta_{M}(\pi_{\tilde{M}_t}^*)-\eta_{M}(\pi_{\tilde{M}_{t+1}}^*)|\\
    &\le |\eta_{M}(\pi_{\tilde{M}_t}^*)-\eta_{\hat{M}_1}(\pi_{\tilde{M}_t}^*)|\\
    &\qquad+|\eta_{\hat{M}_t}(\pi_{\tilde{M}_t}^*)-\eta_{\hat{M}_{t+1}}(\pi_{\tilde{M}_{t+1}}^*)|\\
    &\qquad+|\eta_{M}(\pi_{\tilde{M}_{t+1}}^*)-\eta_{\hat{M}_{t+1}}(\pi_{\tilde{M}_{t+1}}^*)|\\
    &\le(\mathbb{E}_{(s,a)\sim\rho_{\hat{M}_t}^{\pi_{\tilde{M}_t}^*}(s,a)}\delta_{\ell_{1}}(p_{\hat{M}_t}(s,a),p_M(s,a))\\
    &\qquad+\mathbb{E}_{(s,a)\sim\rho_{\hat{M}_{t+1}}^{\pi_{\tilde{M}_{t+1}}^*}(s,a)}\delta_{\ell_{1}}(p_{\hat{M}_{t+1}}(s,a),p_M(s,a)))\frac{1}{2}\gamma V_{max}\\
    &\qquad+\lambda|U_{\hat{M}_t}(\pi_{\tilde{M}_t}^*)-U_{\hat{M}_{t+1}}(\pi_{\tilde{M}_{t+1}}^*)|\\
    &\qquad+D_{\ell_{1}}(p_{\hat{M}_t},p_{\hat{M}_{t+1}})(r_{max}+V_{max})H.
\end{aligned}
\end{equation}

\end{proof}

\end{document}